\documentclass[a4paper, 12pt, one column]{article}

\usepackage[english]{babel}
\usepackage[T1]{fontenc}
\usepackage{booktabs}
\usepackage[skip=7pt, font=small]{caption}
\usepackage{ragged2e}
\usepackage{lscape}
\usepackage{microtype} 
\usepackage{setspace}
\usepackage{parskip}
\usepackage{graphicx}
\usepackage{fontawesome}
\usepackage{array}
\usepackage[utf8]{inputenc}
\usepackage{textcomp}
\usepackage{enumitem}
\usepackage{scrextend}
\usepackage{natbib}

\usepackage[framemethod=TikZ]{mdframed}
\definecolor{lightgray}{rgb}{0.95,0.95,0.95}

\usepackage{tabularx}
\newcolumntype{L}[1]{>{\raggedright\arraybackslash}p{#1}}
\newcolumntype{C}[1]{>{\centering\arraybackslash}p{#1}}
\newcolumntype{R}[1]{>{\raggedleft\arraybackslash}p{#1}}
\usepackage{makecell}
\usepackage{longtable} 
\usepackage{multirow}
\usepackage{multicol}

\usepackage[top=2.5cm, bottom=2.5cm, outer=2.5cm, inner=2.5cm, heightrounded, marginparwidth=1.5cm, marginparsep=0.4cm, margin=2.5cm]{geometry}

\usepackage{graphicx}
\usepackage{hyperref}
\usepackage{amsmath} 
\usepackage{amsfonts} %
\usepackage{amssymb} %
\usepackage{float}
\usepackage{adjustbox}
\usepackage{subfig}
\usepackage{pdfpages}

\hypersetup{
    colorlinks=true,
    linkcolor=blue,
    filecolor=magenta,      
    urlcolor=cyan,
    citecolor=blue,
}

\usepackage{natbib}
\setcitestyle{authoryear,open={(},close={)}}

\usepackage{pgfplots}
\usepackage{tikz}
\pgfplotsset{compat=1.18}
\usepackage{caption}

\usepackage[bottom]{footmisc}

\title{Is GPT-4 a reliable rater? Evaluating Consistency in GPT-4's Text Ratings \thanks{This investigation served as a preliminary study preceding an extensive field study conducted as part of the BMBF-funded DeepWrite project at the University of Passau. The primary objective was to ascertain the consistency of GPT-4's assessments before their integration into authentic scenarios involving students within the realm of Higher Education. We extend our gratitude towards Johann Graf von Lambsdorff, Deborah Voss, and Stephan Geschwind for their contributions in designing the questions, sample solutions, and the field study associated with this investigation.}}
\date{}
\author{
Veronika Hackl\thanks{University of Passau, Faculty of Social and Educational Sciences, Innstrasse 41, 94032 Passau, Germany, Veronika.Hackl@uni-passau.de. Main author.} 
\and
Alexandra Elena Müller\thanks{University of Passau, Faculty of Law, Innstrasse 41, 94032 Passau, Germany. Research assistant.}
\and
Maximilian Sailer\thanks{University of Passau, Faculty of Social and Educational Sciences, Innstrasse 41, 94032 Passau, Germany. Methodology contributor.}
\and
Michael Granitzer\thanks{University of Passau, Faculty of Computer Science and Mathematics, Innstrasse 41, 94032 Passau, Germany. Hypotheses formulation contributor.}
}

\begin{document}
\pagenumbering{gobble}
\maketitle
\abstract{This study investigates the consistency of feedback ratings generated by OpenAI's GPT-4, a state-of-the-art artificial intelligence language model, across multiple iterations, time spans and stylistic variations. The model rated responses to tasks within the Higher Education (HE) subject domain of macroeconomics in terms of their content and style. Statistical analysis was conducted in order to learn more about the interrater reliability, consistency of the ratings across iterations and the correlation between ratings in terms of content and style. The results revealed a high interrater reliability with ICC scores ranging between 0.94 and 0.99 for different timespans, suggesting that GPT-4 is capable of generating consistent ratings across repetitions with a clear prompt. Style and content ratings show a high correlation of 0.87. When applying a non-adequate style the average content ratings remained constant, while style ratings decreased, which indicates that the large language model (LLM) effectively distinguishes between these two criteria during evaluation. The prompt used in this study is furthermore presented and explained. Further research is necessary to assess the robustness and reliability of AI models in various use cases.}


\pagenumbering{arabic}
\section{Introduction}
The integration of AI models, particularly LLMs, into the evaluation of written tasks within educational environments is a burgeoning trend. This trend is driven by the potential of these models to enhance learning outcomes by transforming traditional pedagogical methods.

As the use of these models becomes increasingly pervasive, it is imperative to thoroughly understand and quantify their reliability and consistency. \textit{Elazar et al.} have defined consistency as 'the ability to make consistent decisions in semantically equivalent contexts, reflecting a systematic ability to generalize in the face of language variability' \citep{elazar2021measuring}.

In the context of automated essay grading, inconsistent ratings could lead to unfair outcomes for students, undermining the credibility of the assessment process. Trust in the system 'is highly influenced by users’ perception of the algorithm’s accuracy. After seeing a system err, users’ trust can easily decrease, up to the level where users refuse to rely on a system' \citep[p.3]{Conijn_Kahr_Snijders_2023}. Similarly, in the context of personalized learning, unreliable predictions could result in inappropriate learning recommendations. Therefore, scrutinizing the consistency of AI models is a necessary step towards ensuring the responsible and effective use of these technologies in education \citep{Conijn_Kahr_Snijders_2023}. 

GPT-4, through its emergent Automated Writing Evaluation capabilities, presents a significant advancement in overcoming traditional obstacles inherent in the evaluation of writing tasks. One such obstacle is discourse coherence, a fundamental aspect of writing that refers to the logical and meaningful connection of ideas in a text. In traditional manual grading, assessing discourse coherence can be a subjective and time-consuming process, often leading to inconsistencies in grading. However, GPT-4 with its advanced language understanding capabilities, can analyze the logical flow of ideas in a text, thereby providing a more objective and efficient evaluation of discourse coherence \citep{naismith-etal-2023-automated}.

Feedback plays a crucial role in bridging the gap between a learning objective and the current level of competence and effective feedback, as outlined by {Hattie and Timperley}; encompassing three perspectives: Feed-Back, Feed-Up, and Feed-Forward. Feed-Back involves providing information about the current performance, Feed-Up clarifies the goals, and Feed-Forward gives guidance on how to improve. \citep{HattieTimperley} 'Feedback is a core component of formative assessment processes and has been identified as a powerful factor influencing learning in various instructional contexts, including higher education' \citep{Narciss2020}. Regarding the development of writing skills, feedback on the text plays a crucial role, as it's nearly impossible to improve one's writing abilities without such feedback \citep{schwarze2021}. 

In the context of this study, the AI-generated feedback primarily focuses on the Feed-Back perspective, providing an analysis of the content and style produced by the student. In this scenario of analytic rating, 'the rater assigns a score to each of the dimensions being assessed in the task' \citep{JONSSON2007130}, in our case scores for style and content. The AI-generated feedback in this study is constructed to be adaptive and to assist the learner in figuring out options for improvement. This forms a contrast to non-adaptive or static feedback (e.g. the presentation of a sample solution) which is often used in HE scenarios due to its resource efficiency \citep{SAILER2023101620}. Comprehensive feedback, which includes not only a graded evaluation but also detailed commentary on the students' performance, has been shown to lead 'to higher learning outcomes than simple feedback, particularly in regard to higher order learning outcomes' \citep{vanderkleij2023}. To make the feedback comprehensive and adaptive, it is prompted to include comments on the students' performance as well as numerical ratings and advice on how to improve. 

A key advantage of AI-generated feedback is its immediacy. As noted by Wood and Shirazi (2020), ‘Prompt feedback allows students to confirm whether they have understood a topic or not and helps them to become aware of their learning needs.’\citep[p.~24]{WOOD2020103896}. This immediacy, which is often challenging to achieve in traditional educational settings due to constraints such as class size and instructor workload, can significantly enhance the learning experience by providing students with timely and relevant feedback \citep{haughney2020}. Kortemeyer's observation that 'The system performs best at the extreme ends of the grading spectrum: clearly correct and clearly incorrect solutions are generally reliably recognized [...]' \citep{kortemeyer2023aitool} further underscores the potential of AI models like GPT-4 in assisting human graders. This is particularly relevant in large-scale educational settings where human graders may struggle to consistently identify clearly correct or incorrect solutions due to the sheer volume of work. 

\section{Hypotheses}
 The stability of GPT-4's performance is of significant interest given its potential implications for educational settings where the consistent grading of students' work is paramount. In this investigation, GPT-4 was employed to assess responses to questions within the subject domain of macroeconomics with a focus on both the content and style of the responses. For content, the AI was prompted to evaluate how close the test answer is semantically to the sample solution. A sample solution inserted as demonstration in the prompt allows in-context learning and serves to control the quality of the output \citep{min2022rethinking}. For style, the AI was asked to check whether the language used in the test answer is appropriate for a HE setting and if the response is logically structured and plausible. The questions displayed different levels of complexity. The answers in the test set were created by the authors and subject domain experts, imitating the differing quality of student answers. 

The primary objective of this study is to evaluate the consistency of ratings generated by GPT-4 across multiple iterations, time spans and variations. We demonstrate the agreement between raters and examine various dimensions of consistency. The term raters in our case refers to the different GPT-4 ratings. To provide a comprehensive analysis of GPT-4's performance and application, we propose the following hypotheses:

\begin{enumerate}[leftmargin=2.3cm]
    \item[\textbf{H1:}] The ratings generated by GPT-4 are consistent across multiple iterations.
    
        \item[\textbf{H1.1:}] The ratings generated by GPT-4 are consistent across different time spans, specifically within one week (short-term) and over several months (long-term).
        
        \item[\textbf{H1.2:}] The complexity of the evaluated task does not influence the consistency of GPT-4's ratings.
        
        \item[\textbf{H1.3:}] Different types of feedback (e.g. style, content) do not affect the consistency of GPT-4's performance.
        
    \item[\textbf{H2:}] There is a significant correlation between the ratings for content and style in GPT-4's evaluations. 
    
    \item[\textbf{H3:}] A certain type of prompt framework enables adaptation to new questions and answers, while maintaining consistency in the generated text. 
\end{enumerate}

While each of these hypotheses explores a distinct aspect related to GPT-4's performance and application, collectively, they contribute to a comprehensive and multi-dimensional understanding of GPT-4's potential and limitations in HE.

\section{Methods}

The methods section of this study is designed to provide a comprehensive overview of the research process, detailing the steps taken to address the hypotheses. The research process involves a series of statistical analyses, with the data collection process specifically designed to evaluate the consistency of a LLM in providing feedback and rating students' responses within the subject domain of macroeconomics.

\subsection{Data Collection}

The data collection phase was conducted over a 14-week period from April 2023 to July 2023, with API calls being made at different times and on different days to mimic a realistic usage scenario. The assumption underlying this approach is that the behavior of the model changes over time \citep{chen2023chatgpts}. The API was called through a key within the Audience Response System classEx, which was used to interface with the AI model \citep{giamattei_classex_2019}.

\begin{figure}[H]
\centering
\includegraphics[width=0.8\textwidth]{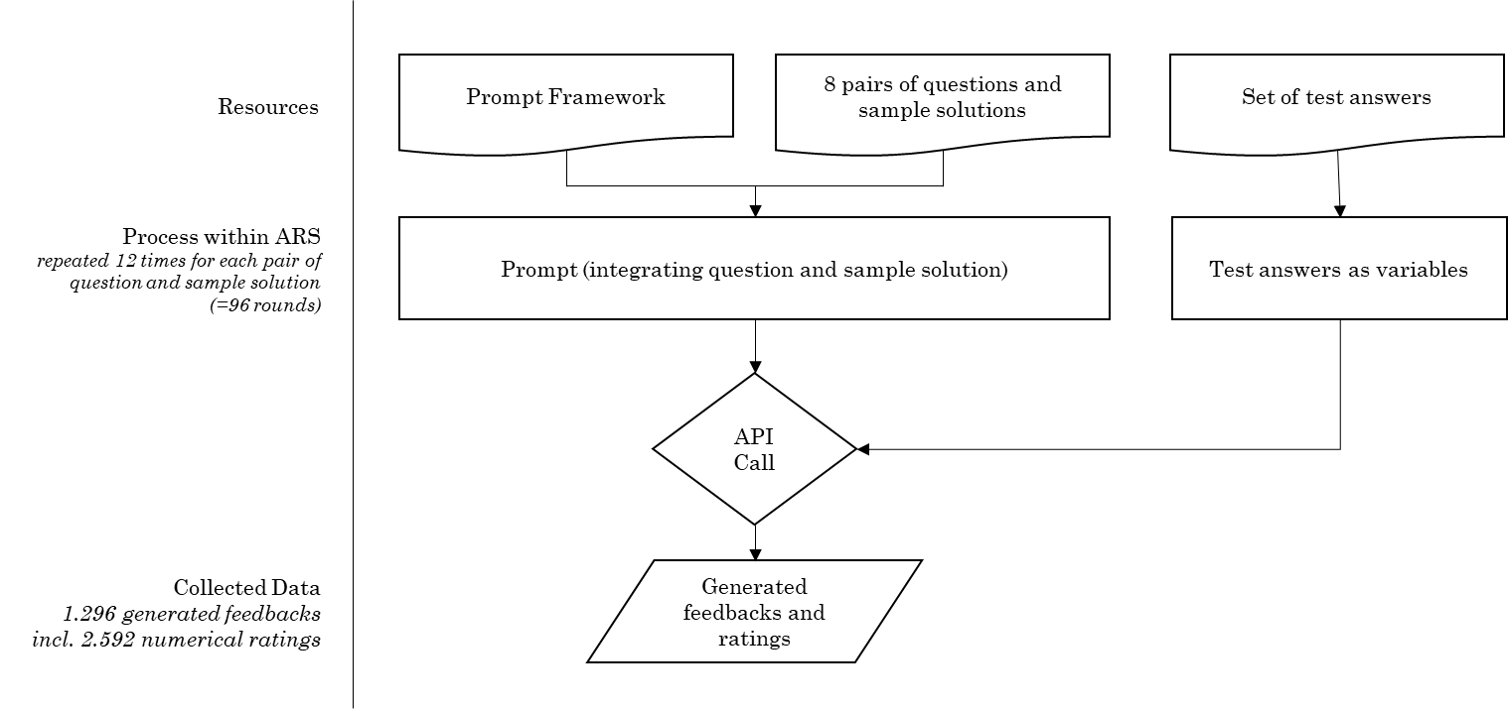}
\caption{Flowchart: Generated texts}
\label{fig:my_label}
\end{figure}

\subsection{Prompt Framework and Test Responses}

The first step in the research process involved the establishment of a prompt framework that serves as a universal structure within the context of this investigation. The goal was to insert new pairs of question and sample solutions without altering the consistency of the output, namely the LLM-generated feedback. Pairs of questions  \citep{Ruth1988DesigningWT}, along with corresponding sample solutions pertinent to macroeconomics, were prepared and integrated into the prompt framework. This integration set the stage for the model to assess students' responses and to generate feedback. First taxonomies aim at structuring prompt formulation approaches. The prompt used in this study would be a Level 4 on the Proposed Prompt Taxonomy TELeR (Turn, Expression, Level of Details, Role) by \citep{santu2023teler}.

\subsubsection{Establishing the prompt framework}

The prompt framework was adapted to ensure consistency in the AI-generated feedback. A tight scaffold or rubric for rating to gain comparable results was used \citep{JONSSON2007130}. The system settings were adjusted to control the randomness of the model's responses, with a temperature setting of 0 used to minimize variability. \citep{si2023prompting,Schulhoff_Learn_Prompting_2022}. By forcing the model into a deterministic behavior, it becomes more consistent in its outputs, while the chances to produce very good or very bad generations decrease. This is a brief documentation of the problems we encountered and the main changes we applied on the path to creating a prompt that works consistently for the use case:

\begin{longtable}{|p{3cm}|p{8.5cm}|}
\hline
\textbf{Problem}  & \textbf{Changes made in prompt} \\
\hline
Output format varies & very clear instructions, ordinal numbers, examples \\
\hline
evaluations not strict enough & role prompting, clear evaluation criteria and application \\
\hline
robustness & shortening the prompt reduces calculation time, fewer outages \\
\hline
multiple identical inputs &  different inputs can be tested at the same time, identical inputs must not be tested in one run as the parameters will then be passed incorrectly and/or the result is homogeneous \\
\hline
informal address with 'Du' & giving a clear instruction in the prompt with example\\
\hline
show star symbols & add the symbol in the prompt \\
\hline
\end{longtable}

\vspace{0.5cm}

\clearpage

This is the final scheme of the prompt framework used for the collection of data (shortened and translated, original language: German): 
\begin{longtable}{|p{3cm}|p{8.5cm}|}
\hline
Element/Function & Prompt Formulation \\
\hline
Role Prompting & You are a professor of macroeconomics and you pose this question to your students: \\
\hline
Variable & <Insert Question here> \\
\hline
Task Description & You evaluate the student's response based on the sample solution using the criteria of content and style, and provide suggestions for improvement. This is the sample solution. It is clearly structured and builds the argument coherently. This solution is both correct in terms of content and very good in terms of style. It would receive 5 out of 5 stars for content and style. Sample solution: \\
\hline
Variable & <Insert sample solution here> \\
\hline
Stepwise Task Description & Please evaluate the student's response based on the sample solution in three steps. \\
\hline
Set Behavior & Here are some general tips for evaluation:
Good feedback is honest and motivating.
Always address the student directly using "you," for example: "Your response." Explain or mention the relevant points you are referring to. \\
\hline
Step 1: Evaluation of content (text feedback) & Step 1: Provide feedback on the content. Answer the following questions: Is the student's response correct in terms of content? Orient yourself to the meaning of the sample solution, but do not mention the sample solution. Are there any areas for improvement? Use a maximum of 2 sentences for this feedback. \\
\hline
Step 2: Evaluation of style (text feedback) & Step 2: Provide feedback on the style: Is the language used by the student appropriate for the field of study? Is the response logically structured and does the argumentation make sense? Are there any areas for improvement? Use a maximum of 2 sentences for this feedback. \\
\hline
Step 3: Evalution (numeric feedback) & Step 3: Evaluate the content and style of the response on a scale of 1 to 5 stars. The rating is based on the feedback on content and style. 1 star indicates a very poor performance. 5 stars indicate a very good performance.
Only display the following for Step 3:
Content: Number of stars (Please also provide the number of stars as a numeral in parentheses)
Style: Number of stars (Please also provide the number of stars as a numeral in parentheses) \\
\hline
Set Format & You provide a concise evaluation divided into 1. to 3.. Always display the stars as follows: star. Output: \\
\hline
\end{longtable}

\subsubsection{Test Responses}

Following the establishment of the prompt framework, test responses were created by domain experts to mimic potential student answers to the given questions. The test set included a variety of responses, ranging from very good answers to nonsense answers and potential prompt injections, to ensure a comprehensive evaluation of the model's performance \citep{liu2023prompt}. An initial set of ten test responses was prepared for the first question. Based on our experience from this initial set, we expanded the test response set to 14 for the subsequent seven questions. We included prompt injections in the test set of answers to fine-tune the prompt and increase the probability to withstand potential attacks \citep{perez2022ignore}.  

\begin{table}[h]
\centering
\begin{tabular}{|l|l|}
\hline
\textbf{Type Number} & \textbf{Type of Test Answer} \\
\hline
1 & copy of sample solution \\
\hline
2 & correct, bullet points \\
\hline
3 & nonsense  \\
\hline
4 & very good answer \\
\hline
5 & opposite of sample solution, but good style \\
\hline
6 & solid answer \\
\hline
7 & incorrect, average style \\
\hline
8 & correct, slightly informal style \\
\hline
9 & average answer \\
\hline
10 & bad content, bad style \\
\hline
11 & only symbols \\
\hline
12 & copy paste of question \\
\hline
13 & prompt injection 1 \\
\hline
14 & prompt injection 2 \\
\hline
\end{tabular}
\caption{Scheme of test set of answers}
\label{tab:icc}
\end{table}

\subsubsection{Format of the generated feedback}

The AI model generated feedback texts for each student response, accompanied by two ratings on a scale of 1 to 5 - one for content ("Inhalt") and one for style ("Stil"). This process was repeated eleven times for each question, resulting in a robust dataset comprising 2.596 generated ratings.

\vspace{0.5cm}
This is the format of the AI-generated feedback:

\vspace{0.5cm}

\definecolor{lightgray}{rgb}{0.95,0.95,0.95}

\begin{mdframed}[backgroundcolor=lightgray]
\begin{quote}
1. Your answer is factually correct and aligns with the essential points of the model solution. You have well explained the impacts of the high population growth rate on the level of the domestic product and its growth.

2. Stylistically, your answer is well-formulated, and the argumentation is logically structured. However, you could explicitly mention the growth model to fully answer the question.

3. Content: \faStar\faStar\faStar\faStar\faStar\ (5) \\
Style: \faStar\faStar\faStar\faStar\ (4)
\end{quote}
\end{mdframed}

\vspace{0.5cm}

\subsubsection{Changing styles}

We further sought to investigate the effects of stylistic changes on the evaluation of content. To serve this goal, the test set of answers was stylistically transformed, keeping the content. Each original student response was rephrased in the style of Warren G., a US-American rapper, with the intention of maintaining the semantic content while changing the stylistic presentation. The corresponding ratings for content and style were again generated through GPT-4. To assess whether the rephrasing affected the content and style ratings, paired t-tests were conducted comparing the original and rephrased ratings for each response. Additionally, the mean and standard deviation of the rating differences were calculated to provide an overview of the magnitude and variation of the changes. This is an illustrative example of an answer out of the original test set (1) and a rephrased version (2). Both texts were translated, the original text is German. The original text, the rephrased text and the corresponding ratings for these examples were as follows: 

\vspace{0.5cm}

\definecolor{lightgray}{rgb}{0.95,0.95,0.95}

\begin{mdframed}[backgroundcolor=lightgray]

\textbf{Original answer, Feedback and Rating} \\
(1) Macroeconomics as engineering starts from human moods and habits. Negative expectations or moods can lead to a decline in domestic production. This is due to fewer expenses being incurred. To reverse this trend, the government should increase its spending. As a result, the domestic product becomes more stable, and no underemployment occurs.
\begin{enumerate}
\item Your answer is substantively correct and covers the essential points of the question. However, you could delve a bit more into the role of the government and why it should act expansively.
\item The style of your answer is appropriate, and the argument is logically structured. However, you could try to elucidate the connections a bit more clearly and formulate the reasons more precisely.
\item Content: \faStar\faStar\faStar\faStar\ (4) \\
Style: \faStar\faStar\faStar\faStar\ (4)
\end{enumerate} 
\end{mdframed}

\vspace{0.5cm}

\definecolor{lightgray}{rgb}{0.95,0.95,0.95}

\begin{mdframed}[backgroundcolor=lightgray]
\textbf{Rephrased answer, Feedback and Rating} \\
(2) Macroeconomics, baby, is like fine-tuning an engine block - it's all about the vibes and routine of the people. If folks have a negative vibe or their attitude isn't positive, it can delay the economic performance, the GDP, because they're simply spending less dough. But you know what, the government can change the game if they themselves start splashing more cash. This keeps the GDP in a stable zone, and we don't have people who have to sit at home twiddling their thumbs.
\begin{enumerate}
\item Your answer is substantively correct as you highlight the impacts of moods and expectations on domestic production and the role of government spending. However, you could define and explain the terms "domestic product" and "underemployment" more precisely.
\item Your style is informal and could be considered inappropriate in an academic context. Try to use a more formal language and structure your argument more clearly.
\item 3. Content: \faStar\faStar\faStar\ (3) \\
Style: \faStar\faStar\ (2)
\end{enumerate}
\end{mdframed}

\subsection{Statistical Analysis}

\subsubsection{Intraclass Correlation Coefficient}

The Intraclass Correlation Coefficient (ICC) is a statistical measure in order to assess the level of agreement or consistency among raters. A perfect ICC score of 1 indicates perfect agreement or consistency among raters, while a score of 0 indicates no agreement nor consistency. ICC estimates and their 95\% confident intervals were calculated using RStudio based on a mean-rating, absolute-agreement, 2-way mixed-effects model. For the decision which ICC calculation to use the flowchart proposed by Koo et.al was used. The type of the reliability study is 'inter-rater reliability'. We assign the different iterations of GPT-4 the role of different raters and assume that the same set of raters (GPT-4 at different points of time) rates all subjects. The model chosen is the two-way mixed effects model as we assume to have a specific sample of raters. The model type decided for is based on the mean of multiple raters. Both the model definitions “absolute agreement” and “consistency” were chosen. This results in the 2-way mixed-effects model. The caveat in the ICC model chosen in the analysis is that it only represents the reliability of the specific raters involved in this experiment \citep{KOO2016155}. As Generative AI remains a "black box" system, this was considered to be the most suitable model \citep{cao2023comprehensive}.

The extracted numerical ratings from the feedback texts formed the dataset for the statistical analyses and were utilized to calculate the ICC, providing a measure of the consistency of the ratings generated by the AI model. 

\subsubsection{Correlation Analysis and Rating Differences}

In order to answer H2, a correlation analysis was conducted. This analysis involved calculating the correlation coefficient between the content and style ratings generated by the AI model. The correlation coefficient provides a measure of the strength and direction of the relationship between the content and style ratings, thereby providing insights into the model's grading criteria. Skewness is a measure of the asymmetry of the probability distribution of a real-valued random variable about its mean. In this study, the skewness of the rating distributions was calculated to examine the symmetry of the data. The purpose of this analysis was to evaluate the extent to which the ratings deviated from a normal distribution. 

\section{Results}

The results section of this study presents the findings from the statistical analyses conducted to address the hypotheses. The analyses include the computation of Intraclass Correlation Coefficients (ICCs), skewness measures for content and style ratings, and a correlation analysis between content and style ratings.

\subsection{Intraclass Correlation Coefficients}

The tables 4 and 5 present the ICCs for ratings on content (Inh) and style (Stil) based on two different measurements. Table 4 reports ICCs from the initial ten measurements conducted between April and June 2023. The ICC values for both absolute agreement and consistency for content and style are extremely high (0.999), suggesting almost perfect agreement and consistency among raters. The 95\% confidence intervals (CI) are as well tight, ranging from 0.998 to 0.999, indicating if the study was replicated, the true ICC would be expected to fall within this range 95\% of the time. The F-tests are significant (p < 0.001), providing statistical evidence that the raters are reliably consistent and in agreement with each other in their ratings.

Table 5 reports ICCs from a control measurement. The ratings were obtained from two raters: the first being an average rating compiled from ten raters across the months of April to June, and the second being a single rater evaluating in July. The result shows lower ICC values of 0.944 for both Inh and Stil. While these are still high values indicating good agreement, they are not as high as the ICC values in table 4. This implies that while a robust agreement persists between the mean rating and the July rater, it is not as pronounced as the concordance among the ten raters. This inference suggests a temporal evolution in the model's behavior, necessitating diligent continuous assessment for its utilization in educational tasks.

\begin{table}[h]
\centering
\begin{tabular}{|l|l|l|l|}
\hline
\textbf{ICC Type} & \textbf{ICC Value} & \textbf{95\% CI} & \textbf{F-Test} \\
\hline
Absolute agreement (Inh) & 0.999 & 0.999 - 0.999 & F(107,971) = 1332, p < 0.001 \\
\hline
Absolute agreement (Stil) & 0.999 & 0.998 - 0.999 & F(107,971) = 689, p < 0.001 \\
\hline
Consistency (Inh) & 0.999 & 0.999 - 0.999 & F(107,963) = 1332, p < 0.001 \\
\hline
Consistency (Stil) & 0.999 & 0.998 - 0.999 & F(107,963) = 689, p < 0.001 \\
\hline
\end{tabular}
\caption{Reporting of Intraclass Correlation Coefficients (ICC)}
\label{tab:icc}
\end{table}

\begin{table}[h]
\centering
\begin{tabular}{|l|l|l|l|}
\hline
\textbf{ICC Type} & \textbf{ICC Value} & \textbf{95\% CI} & \textbf{F-Test} \\
\hline
Absolute agreement (Inh) & 0.944 & 0.918 - 0.962 & F(107,108) = 17.8, p < 0.001 \\
\hline
Absolute agreement (Stil) & 0.944 & 0.918 - 0.962 & F(107,108) = 17.8, p < 0.001 \\
\hline
Consistency (Inh) & 0.944 & 0.918 - 0.962 & F(107,107) = 17.8, p < 0.001 \\
\hline
Consistency (Stil) & 0.944 & 0.918 - 0.962 & F(107,107) = 17.8, p < 0.001 \\
\hline
\end{tabular}
\caption{Reporting of Intraclass Correlation Coefficients (ICC) (mean rating of 10 raters from April to June, contrast rating of July)}
\label{tab:icc}
\end{table}

\subsection{Correlation between Content and Style Ratings}
The relationship between the average content (Inh) and style (Stil) ratings was examined to assess the interplay between these two dimensions of evaluation. A correlation analysis was conducted, yielding a correlation coefficient of 0.87. This high value indicates a strong positive relationship between content and style ratings, suggesting that responses rated highly in terms of content were also likely to receive high style ratings, and vice versa.

This strong correlation underscores the interconnectedness of content and style in the evaluation process, suggesting that the AI model does not distinctly separate these two aspects but  rather views them as interrelated components of a response's overall quality. When the student answers were rephrased in a different style, we found that the average difference in content ratings before and after rephrasing was approximately 0.056 (stars rating), with a standard deviation of around 1.33. The paired t-test revealed no significant difference in content ratings between the original and rephrased responses (t = 0.434, p = 0.665). In terms of style ratings, the average difference before and after rephrasing was approximately 0.241, with a standard deviation of around 1.37. The paired t-test suggested a marginally significant difference between the original and rephrased style ratings (t = 1.813, p = 0.073).

The skewness of the content and style ratings was calculated to assess the distribution of these ratings. A positive skewness value indicates right-skewness, while a negative value indicates left-skewness. In this study, the positive skewness values for content suggest that the AI model tended to give higher ratings for content (see Table 6). Conversely, the majority negative skewness values for style suggest a left-skewness, indicating that the model was more critical in its ratings for style (see Table 7).

These skewness values provide insights into the AI model's rating tendencies. The right-skewness for content ratings suggests that the AI model may be more lenient in its content evaluations or that the student responses were generally of high quality. The left-skewness for style ratings, on the other hand, suggests that the AI model may have stricter criteria for style or that the style of the student responses varied more widely. These insights can inform future refinements of the AI model to ensure more balanced and fair evaluations.

\begin{table}[ht]
\centering
\begin{minipage}{.5\linewidth}
\centering
\begin{tabular}{|c|c|}
\hline
Rater & Skewness \\
\hline
1\_Inh & 0.107009 \\
2\_Inh & 0.080385 \\
3\_Inh & 0.094007 \\
4\_Inh & 0.116521 \\
5\_Inh & 0.076956 \\
6\_Inh & 0.096934 \\
7\_Inh & 0.126752 \\
8\_Inh & 0.089091 \\
9\_Inh & 0.094007 \\
10\_Inh & 0.090488 \\
11\_Inh & 0.299014 \\
\hline
\end{tabular}
\captionof{table}{Skewness for Content Ratings}
\label{table:skewness content}
\end{minipage}%
\begin{minipage}{.5\linewidth}
\centering
\begin{tabular}{|c|c|}
\hline
Rater & Skewness \\
\hline
1\_Stil & -0.037198 \\
2\_Stil & -0.043986 \\
3\_Stil & 0.029177 \\
4\_Stil & -0.017839 \\
5\_Stil & -0.047688 \\
6\_Stil & 0.000873 \\
7\_Stil & -0.040248 \\
8\_Stil & -0.050956 \\
9\_Stil & -0.013981 \\
10\_Stil & -0.017839 \\
11\_Stil & -0.147365 \\
\hline
\end{tabular}
\captionof{table}{Skewness for Style Ratings}
\label{table:skewness style}
\end{minipage}
\end{table}

\section{Discussion}
The findings of this study provide insights into the potential of AI models, specifically GPT-4, in evaluating student responses in the context of macroeconomics. 

\begin{itemize}
\item The high ICC values for both content and style ratings suggest that the AI model was able to consistently apply well-defined evaluation criteria at different points of time and with variations of style and content. 
\item The ICC values were lower when calculated with another set of feedbacks generated after a timespan of several weeks. 
\item The high level of concurrence between ratings underlines the dependability of the evaluation method employed in this study. 
\item The positive correlation between content and style ratings underscores the interconnectedness of content and style in the evaluation process. 
\item Rephrasing the answers stylistically did not significantly affect the content ratings, implying that GPT-4 was able to separate content from style in its evaluations. 
\item The ICC values show that forcing GPT-4 into a deterministic behavior through prompt- and system settings works.
\end{itemize}

It is important to note the limitations of AI models as their application in educational settings is not free of challenges. As stated in this paper, the ICC values differ for ratings at different points of time. There are variations in consistency for different levels of question complexity. Other limitations are being mentioned in OpenAI's technical report on GPT-4: AI models can sometimes make up facts, double-down on incorrect information, and perform tasks incorrectly \citep{openai2023gpt4}. Another challenge is the "black box" problem, as discussed by \citep{cao2023comprehensive}. This refers to the lack of transparency and interpretability of AI models, which can hinder their effective use in educational settings. Further research is needed to address this issue and enhance the transparency and interpretability of AI models.

Despite these challenges, there are promising avenues for enhancing the capabilities of AI models. The provision of feedback to macroeconomics students can be characterized as an emergent capability of the AI model. Emergence is a phenomenon wherein quantitative modifications within a system culminate in qualitative alterations in its behavior. This suggests that larger-scale models may exhibit abilities that smaller-scale models do not, as suggested by \citep{wei2022emergent}. However, a direct comparison with GPT-3.5 is needed to substantiate this claim. The potential of AI models in providing feedback can be further enhanced by improving their "Theory of Mind" or human reasoning capabilities, as suggested by \citep{moghaddam2023boosting}. This could lead to more nuanced and contextually appropriate feedback, thereby enhancing the learning experience of students. Furthermore, the study by \citep{Fu2023GPTScoreEA} points to the potential of using AI models to audit generative AI. This opens up new avenues for ensuring the quality and reliability of AI-generated content. Above that, the use of smaller models should be encouraged \citep{bursztyn-etal-2022-learning} as well as the idea to evaluate AI-generated feedbacks either by a human rater or an AI before shown to the student \citep{perez2022discovering}. 

In conclusion, while the results of this study are encouraging, they underscore the need for further research to fully harness the potential of AI models in educational settings. Future studies should focus on addressing the long-term performance, but also the limitations of AI models and exploring ways to enhance their reliability, transparency, and interpretability.  

\clearpage

\bibliographystyle{plain}

\end{document}